\pdfoutput=1

\documentclass[11pt]{article}

\usepackage[final]{acl}

\usepackage{times}
\usepackage{latexsym}

\usepackage[T1]{fontenc}

\usepackage[utf8]{inputenc}

\usepackage{microtype}

\usepackage{inconsolata}

\usepackage{graphicx}
\usepackage{float}
\usepackage{multirow}
\usepackage[most]{tcolorbox}
\tcbuselibrary{listings}
\usepackage{amsmath} 
 
\title{GIIFT: Graph-guided Inductive Image-free\\Multimodal Machine Translation}

\author{Jiafeng Xiong \\
  Department of Computer Science\\
  University of Manchester\\
  \texttt{jiafeng.xiong@manchester.ac.uk}\\\And
   \quad\qquad  Yuting Zhao\\
  \quad\qquad Department of Advanced Information Technology \\
  \quad\qquad  Kyushu University \\
  \quad\qquad  \texttt{zhao.yuting.095@m.kyushu-u.ac.jp} \\}

\begin{document}
\maketitle
\begin{abstract}
Multimodal Machine Translation (MMT) has demonstrated the significant help of visual information in machine translation. However, existing MMT methods face challenges in leveraging the modality gap by enforcing rigid visual-linguistic alignment whilst being confined to inference within their trained multimodal domains. In this work, we construct novel multimodal scene graphs to preserve and integrate modality-specific information and introduce \textbf{GIIFT}, a two-stage \textbf{\underline{G}}raph-guided \textbf{\underline{I}}nductive \textbf{\underline{I}}mage-\underline{\textbf{F}}ree MM\underline{\textbf{T}} framework that uses a cross-modal Graph Attention Network adapter to learn multimodal knowledge in a unified fused space and inductively generalize it to broader image-free translation domains. Experimental results on the Multi30K dataset of English-to-French, English-to-German, and English-to-Ukraine tasks demonstrate that our GIIFT surpasses existing MMT baselines even without images during inference.
Results on the WMT benchmark show significant improvements over the image-free MMT translation baselines, demonstrating the strength of GIIFT towards inductive image-free inference~\footnote{Code is available at: \url{https://github.com/xjiaf/GIIFT}}.
\end{abstract}

\section{Introduction}

Multimodal machine translation (MMT) aims to improve traditional text-only neural machine translation (NMT) by incorporating multimodal data, particularly visual inputs~\cite{specia_shared_2016,elliott-etal-2017-findings,barrault-etal-2018-findings}.  
Existing methods mostly focus on forcing the alignment between image and text to improve MMT, that they have demonstrated the effectiveness benefited from the aligned visual information in disambiguating text~\citep{ive_distilling_2019,wra_papers,futeral_tackling_2023}.
However, an aligned form of multimodal dataset in both the training and inference phases has disabled the MMT model to generalize further in the normal pure-text machine translation setup. 
Although the rich information of images can bring benefits to translation beyond the level of text, when aligned image information is indispensable for inference in the translation process, the application of MMT models will be severely limited.
Therefore, the inability to get rid of aligned images in the inference phase is the critical bottleneck for the flexible application of MMT models.

To address the bottleneck mentioned above, there have been a few methods that attempt to explore resolutions for achieving image-free inference in MMT models.
For example, synthesizing visual hallucination from text or textual scene graph during training is used for the image-free inference~\citep{li_valhalla_2022,fei_scene_2023}; transfer learning from the image-to-text captioning task to the text-to-text translation task~\citep{gupta_cliptrans_2023}. 
However, none of these efforts has managed to consistently reach the performance of fully bridging the gap between multimodal and image-free inference.
There are still critical challenges in advancing image-free inference in MMT.

\begin{figure}[!t]
    \centering
    \includegraphics[width=1.0\linewidth]{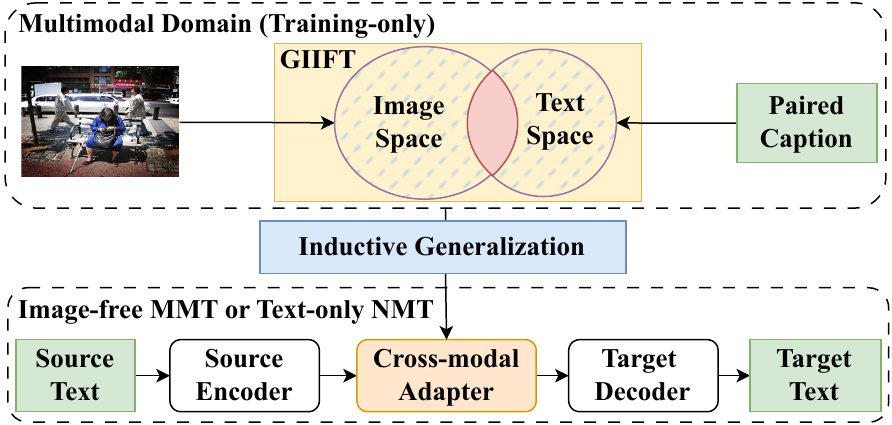}
\caption{The inductive image-free generalization of GIIFT. GIIFT inductively learns from an entire multimodal domain, then enables image-free inference for MMT or text-only NMT via cross-modal generalization. In contrast, previous models can only learn from the limited overlap between image and text in red.}
    \label{fig:image-free MMT}
\end{figure}

First, previous models learn inadequately because they forcibly align the modality gap, typically the intrinsic information imbalance between images and texts~\citep{schrodi2025effectstriggermodalitygap}. 
Consider the case in Figure~\ref{fig:image-free MMT}, this constraint limits image-free inference generalized from only the red overlap in the multimodal domain. Recent work~\citep{10656738,schrodi2025effectstriggermodalitygap}, however, shows that accepting this gap and exploiting the full information (the entire collection in Figure~\ref{fig:image-free MMT}) substantially enhances multimodal learning. 
Second, current image-free MMT approaches fail to realize cross-modal generalization, resulting in a marked drop in image-free inference compared with traditional MMT with multimodal inference~\cite{fei_scene_2023}.
Third, current MMT models are transductive~\cite{sutskever_sequence_2014}, which struggle with inductively generalizing from multimodal domains to text-only domains for broader applications.
Facing these challenges, we investigate the following research questions in this work: \textit{RQ1: How can we fully represent different modalities to embrace the modality gap? RQ2: Can we effectively make image-free inference via cross-modal generalization without downgrading performance? RQ3: Can MMT have the inductive ability to generalize multimodal domains to broader text-only domains?}

We propose \textbf{GIIFT}, a two-stage \textbf{\underline{G}}raph-guided \textbf{\underline{I}}nductive \textbf{\underline{I}}mage-\underline{\textbf{F}}ree MM\underline{\textbf{T}} framework.
As shown in Figure~\ref{fig:image-free MMT}, the GIIFT learns from the entire multimodal domain in the first stage, and aims to achieve inductive generalization for image-free MMT or text-only NMT in the second stage.
Graph-structured novel Multimodal Scene Graph (MSG) and Linguistic Scene Graph (LSG) are proposed to represent the multimodal domain, in which each modality informs and enriches the other by graph representation in the unified space. 
Specifically, we extract visual relationships from images and linguistic relationships from text, then preserve and integrate them via global super nodes to construct MSGs. LSG is a linguistic version, preserving linguistic relationships with global super nodes. These relations' and nodes' features are mutually enriched and uniformly initialized via M-CLIP~\cite{carlsson_cross-lingual_2022}. 
To enable cross-modal generalization to image-free or broader text-only domains, GIIFT is designed with a cross-modal GAT adapter to inductively learn multimodal knowledge from the multimodal domain via MSGs in the first stage, and generalize it for image-free inference via LSGs in the second stage, based on a replaceable backbone, mBART~\cite{liu_multilingual_2020}.

\textbf{Overall, the main contributions are:}\\
\textbf{(i)} We build novel MSGs and LSGs to fully represent different modalities in a unified space for embracing the modality gap.\\
\textbf{(ii)} We propose the two-stage GIIFT framework with a novel lightweight GAT adapter to achieve inductive cross-modal generalization for image-free inference MMT via MSGs and LSGs.\\
\textbf{(iii)} Experimental results on En$\rightarrow$\{Fr, De, Uk\} Multi30K show that GIIFT outperforms most of the existing MMT methods even without image during inference. 
On En$\rightarrow$\{Fr, De\} WMT, GIIFT surpasses the best image-free baseline, CLIPTrans, by average gains of \textbf{+1.92~(8.00\%) BLEU} and \textbf{+2.82~(4.80\%) METEOR}, demonstrating effective induction from multimodal Multi30K to other text-only NMT domains.\\
\textbf{(iv)} Further analysis demonstrates that the proposed GIIFT can effectively embrace modality gaps via MSGs and achieve robust image-free inference via two-stage cross-modal generalization, and shows that the GIIFT can achieve consistent performance of fully bridging the gap between multimodal inference and image-free inference.

\section{Related Work}
\subsection{Multimodal Machine Translation}

MMT research integrates visual and textual information for machine translation with an increasing number of models~\cite{gronroos_memad_2018,huang_unsupervised_2020,ra_papers,cheng_soul-mix_2024}. Early approaches commonly adopted RNN-based architectures enhanced with attention mechanisms to incorporate global or spatial visual features~\cite{calixto_doubly-attentive_2017}. Transformer variants soon supplanted RNNs, introducing tighter cross-modal fusion such as dynamic token re-weighting~\cite{caglayan_lium-cvc_2018,lin_dynamic_2020}, double attention mechanisms~\cite{zhao-etal-2020-double}, gating mechanisms~\cite{wu_good_2021}, multimodal adapters~\cite{Zhao2022MultimodalRF} or multi-granular fusion via graph structures~\cite{krishna_visual_2017,wang_scene_2018,yin_novel_2020}. Recent research leverages pre-trained resources such as CLIP~\cite{gupta_cliptrans_2023,li_valhalla_2022} or BERT~\cite{li_what_2020}. Within the widely adopted encoder-decoder framework, MMT research has progressed along two fronts in representation and inference:\\
\noindent\textbf{(1) Visual-linguistic representation.}
Most MMT models enforce rigid visual-linguistic alignment. Disambiguation work~\cite{ive_distilling_2019,wra_papers,futeral_tackling_2023} links each textual token to a matching image region to resolve lexical ambiguity. 
UMMT\cite{fei_scene_2023} aligns every hallucinated visual scene graph node with textual counterpart. CLIPTrans~\cite{gupta_cliptrans_2023} trains sequentially on two stages, image captioning and the corresponding translation, enforcing alignment across both stages. Such alignments discard modality-specific information by merely learning the overlap between modalities.

\noindent\textbf{(2) Image-free inference.}
Previous MMT relies on access to the paired image at test time. To mitigate this limitation, researchers have pursued three lines of work. First, retrieval-based models replace the missing picture with visual features fetched from an indexed caption-image bank into the decoder~\cite{zhang_neural_2020}. Second, hallucination methods~\cite{johnson_image_2018,li_valhalla_2022,fei_scene_2023} synthesize visual inputs from texts. Third, using transfer learning to train the NMT models with images~\cite{gupta_cliptrans_2023}.

\subsection{Graph Neural Networks}

GNN is powerful in modeling relational structures by leveraging message-passing mechanisms~\cite{wu_comprehensive_2020,liang_survey_2022}, which iteratively aggregates and updates nodes' representation with information from their neighbors, capturing both local and global relational patterns. Some GNNs, such as Graph Convolutional Network (GCN)~\cite{kipf_semi-supervised_2017} and its variants, are only transductive, while others, including GAT~\cite{velickovic_graph_2018}, and GraphSAGE~\cite{hamilton_inductive_2017}, 
also enable inductive learning~\citep{battaglia_relational_2018,xiong_survey_2025} to handle previously unseen nodes~\cite{zhou_graph_2020}. GNNs also use hierarchical or global pooling techniques~\cite{ying_hierarchical_2018,lee_self-attention_2019,gao_graph_2019} to capture subgraph-level or graph-level embeddings~\cite{zhou_graph_2020}. 

\section{Methodology}
In this work, we construct unified MSGs and LSGs to generalize multimodal knowledge for image-free inference. We design the GIIFT framework with two-stage continuous learning via a lightweight GAT adapter for inductive cross-modal generalization.
Our methodology is structured as follows: 1) We introduce the problem definition of our inductive image-free inference (Subsection~\ref{sec:problem}). 2) The details of the MSG and LSG scene graph generation (Subsection~\ref{sec:scene graph}). 3) The description of the GIIFT framework (Subsection~\ref{sec:giift}).

\subsection{Problem Definition}\label{sec:problem}
Let $\mathcal{D}_m$ be a multimodal multilingual dataset of triplets $(i, c_s, c_t)$, where $i$ is an image and $(c_s,c_t)$ are its source and target captions, and let $\mathcal{D}_l$ be a text-only parallel corpus of pairs $(t_s,t_t)$. Traditional MMT methods align $i$ with $c_s$ during training and then perform image-free inference based on that alignment, but the visual knowledge remains tied to $c_s$ and $\mathcal{D}_m$. Our inductive image-free inference instead learns multimodal knowledge from $i$ and $(c_s,c_t)$ in $\mathcal{D}_m$ that cross-modally generalizes to both $\mathcal{D}_m$ or translation pairs $(t_s,t_t)\in\mathcal{D}_l$.

\subsection{Scene Graph Generation}\label{sec:scene graph}
\begin{figure}[htbp]
    \centering
    \includegraphics[width=1\linewidth]{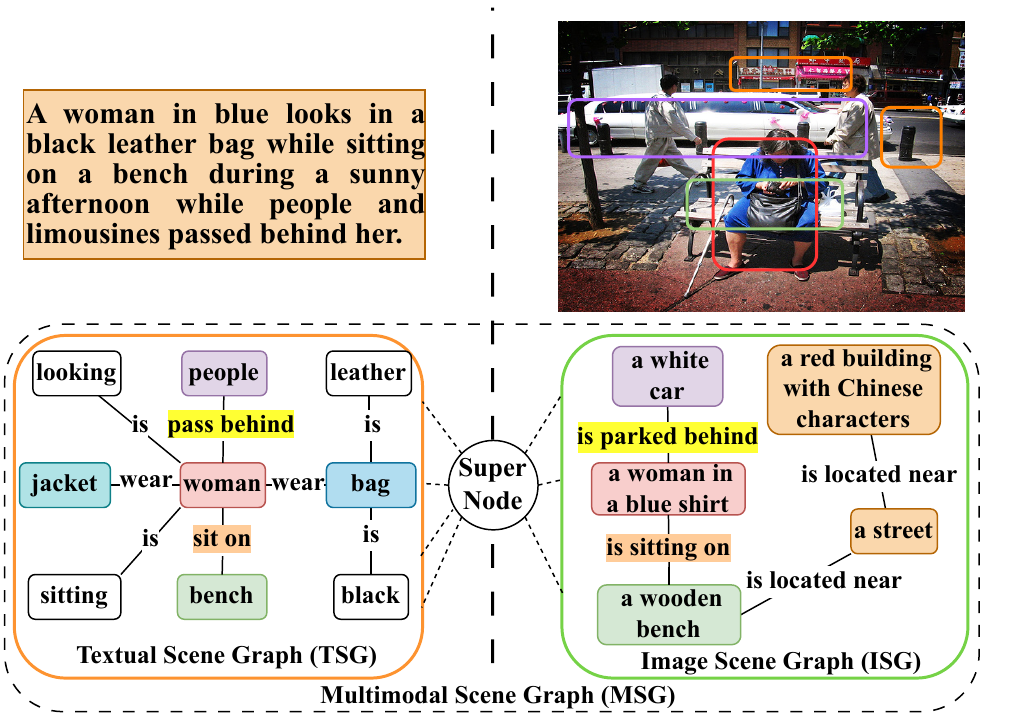}
    \caption{Representation of textual and visual information via MSG. Corresponding objects, attributes, and relations across both Textual Scene Graphs and Image Scene Graphs are depicted using identical colors.}
    \label{fig:scene graph}
\end{figure}
\noindent To preserve modality-specific information and extract complex relations (e.g., spatial relations, environmental scene, and event states of people and objects) during data preprocessing, we use a multimodal Large Language Model~(LLM) as the image parser and an off-the-shelf text parser to build the MSG and LSG, respectively. Figure~\ref{fig:scene graph} shows that the MSG comprises an Image Scene Graph~(ISG) and a Textual Scene Graph~(TSG), and a visual super node to bridge ISG and TSG included in a unified space, whereas the LSG replaces the visual super node with a textual one and omits the ISG.

\begin{figure*}[ht]
    \centering
    \includegraphics[width=1\linewidth]{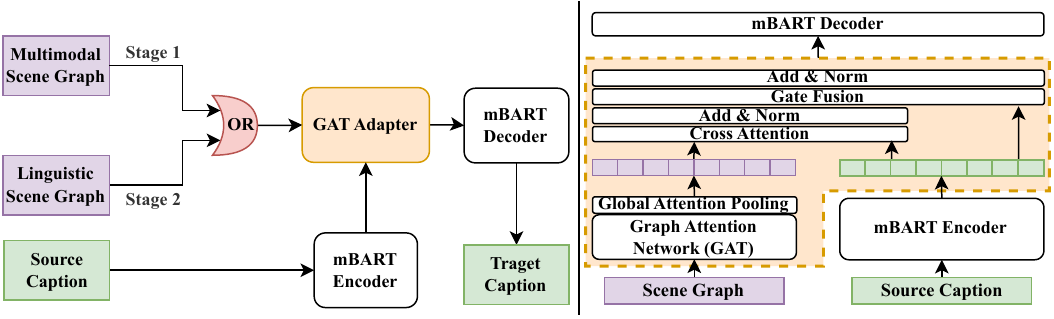}
    \caption{Left: Overview of the two-stage GIIFT framework. Stage~1: multimodal learning via MSGs. Stage~2: cross-modal generalization via LSGs. Right: Overview of the architecture of the cross-modal GAT adapter, which inductively learns and fuses the multimodal knowledge for the backbone, mBART.}
    \label{fig:framework}
\end{figure*}

\noindent\textbf{(1) Image Scene Graph.}
ISGs are obtained from images $i$ using LLaVA-34B~\cite{liu_visual_2023}. To obtain coherent and well-structured outputs, we set the sampling temperature to 0 for deterministic generations and raise it to 0.4 for more challenging cases requiring exploratory outputs. Our prompts (details in Appendix~\ref{sec:prompt}) comprise: 
\textbf{(i) Task Description}, specifying how to formulate relations and produce structured scene graph content; 
\textbf{(ii) Negative Examples}, illustrating common errors in output formatting and how to rectify them; 
\textbf{(iii) Format Examples}, providing abstract but well-formed scene graph templates without using specific objects, thus preventing prompt contamination. Therefore, we generate ISGs with unique and relational visual information, such as event states.\\
\noindent\textbf{(2) Textual Scene Graph.}
TSGs are parsed by FACTUAL~\citep{li_factual_2023} from texts $c_s$ or $t_s$. FACTUAL is lighter and more efficient than LLMs for large-scale corpora. TSGs encode entities and their relations as the textual analogue to ISGs. Thus, we obtain TSGs with structured linguistic relationships and unique semantic information.\\
\noindent\textbf{(3) Multimodal Scene Graph.} 
For each pair $(i, c_s)$, we merge the ISG and TSG into an MSG by introducing a super node that encodes holistic image embeddings from the M-CLIP image encoder. This super node connects to all ordinary ISG and TSG nodes to unite modality-specific information and deliver diverse granular information, serving as a critical bridge to accept the modality gap and build multimodal relationships. We embed all ordinary node and relation features by the M-CLIP text encoder, enabling a unified representation of multimodal relationships and inductive foundations.\\
\noindent\textbf{(4) Linguistic Scene Graph.} 
For cross-modal generalization, we construct the LSG for text pairs $(t_s, t_t)$ by retaining only the TSG with a super node representing the entire text embedding via the M-CLIP text encoder, which shares the unified hidden space with MSG's. Similarly, we embed all ordinary nodes and edges using the M-CLIP text encoder. The super node connects to all ordinary TSG nodes, enabling multi-granular textual information for image-free inference.

\subsection{GIIFT Framework}\label{sec:giift}

Harnessing scene graphs as modality-bridges, GIIFT uses MSGs and LSGs, which are inductively learned for multimodal knowledge and generalized for image-free translation, respectively, via an essential cross-modal GAT adapter to guide the backbone mBART in two-stage training pipelines. The proposed lightweight GAT adapter is completely decoupled from the pre-trained mBART backbone (as shown in Figure~\ref{fig:framework}), which can be flexibly incorporated with other language models for deployment. 

\subsubsection{Two-stage Training Framework}

\noindent\textbf{Stage~1: Multimodal Learning via MSG.}  As shown in Figure~\ref{fig:framework}~(Left), Stage~1 trains a shared GAT adapter on MSGs from paired images and captions, inducing multimodal knowledge that guides image-free translation. Within the MSG, \(\mathcal{N}_i^{\mathrm{SG}}\) is the neighbors of the ordinary node \(i\) from ISG or TSG, reflecting the structured relationships from image or text. The embedding \(\mathbf{Z}_i^{(l)}\) of node $i$ at layer \(l, (0 \leq l \leq L)\) in GAT is calculated recursively as:
\begin{equation}
\begin{split}
\mathbf{Z}_i^{(l)}=\sigma(
\!\!\!\sum_{j\in\mathcal{N}_i^{\mathrm{SG}}}\!\!\!\!\alpha_{ij}^{(l)}\,\phi(\mathbf{W}[\mathbf{Z}_i^{(l-1)}\|\mathbf{Z}_j^{(l-1)}\|\mathbf{E}_{ij}])\\
+\alpha_{i,\mathrm{SN}}^{(l)}\,\phi(\mathbf{W}[\mathbf{Z}_i^{(l-1)}\|\mathbf{Z}_{\mathrm{SN}}^{(l-1)}\|\mathbf{E}_{i,\mathrm{SN}}])
).
\end{split}
\label{eq:ordinary_node_msg_l}
\end{equation}
Here, \(\mathbf{Z}_i^{(l-1)}\) is the embedding of node \(i\) at the previous layer (with initial node embedding \(\mathbf{Z}_i^{(0)}\) via M-CLIP text encoder), \(\mathbf{Z}_{\mathrm{SN}}^{(l-1)}\) is the global multimodal embedding from the super node at layer \(l-1\) which is passed to all the ordinary nodes, and \(\mathbf{E}_{ij}\) or \(\mathbf{E}_{i,\mathrm{SN}}\) denotes edge embedding of the relationships in the scene graph via M-CLIP text encoder, the \(\alpha_{ij}^{(l)}\) and \(\alpha_{i,\mathrm{SN}}^{(l)}\) are attention weights, \(\sigma(\cdot)\) is an activate function and $\phi(\cdot)$ is the LeakyReLU. \(\mathbf{E}_{ij}\) is the embedding of the relation content in the scene graph, obtained with the M‑CLIP text encoder, whereas \(\mathbf{E}_{i,\mathrm{SN}}\) is the embedding of the edge between the super‑node and each ordinary node $i$. Because these super‑node edges have no textual content, we initialize \(\mathbf{E}_{i,\mathrm{SN}}\) as a ones vector whose dimensionality matches that of the embeddings produced by the M‑CLIP text encoder.

From Eq.~\eqref{eq:ordinary_node_msg_l}, we can observe that the shared weight \(\mathbf{W}\) enables learning of multi-granular multimodal relationships by jointly processing local textual embeddings from ISG or TSG nodes (\(\mathcal{N}_i^{\mathrm{SG}}\)) and global multimodal context from the super node. This shared weight \(\mathbf{W}\) is crucial in capturing these multimodal relationships and will be leveraged for the cross-modal generalization process in Stage~2.

The initial MSG super node provides the global image embedding  $\mathbf{Z}_{\mathrm{SN}}^{\mathrm{MSG}(0)}$ and aggregates information from all ordinary nodes as follows:
\begin{equation}
\mathbf{Z}_{\mathrm{SN}}^{\mathrm{MSG}(l)}\!\!=\!\sigma(\!\!\!\!\!\sum_{i\in\mathcal{N}_{\mathrm{SN}}^{\mathrm{MSG}}}\!\!\!\!\alpha_{\mathrm{SN},i}^{(l)}\phi(\mathbf{W}[\mathbf{Z}_{\mathrm{SN}}^{(l-1)}\|\mathbf{Z}_i^{(l-1)}\|\mathbf{E}_{\mathrm{SN},i}]))
\label{eq:supernode_msg_l}
\end{equation}
where \(\mathcal{N}_{\mathrm{SN}}^{\mathrm{MSG}}\)  contains all ordinary nodes in MSG.

Through the Global Attention Pooling~\cite{li_gated_2015} layer in the GAT adapter, we then obtain the multimodal graph representation \(\mathbf{Z}_g^\text{MSG}\in\mathcal{D}_m\):
\begin{equation}
\mathbf{Z}_g^{\mathrm{MSG}} = \text{AttnPool}(\{\mathbf{Z}_i^{\mathrm{MSG}}: i\in\mathcal{V}_{\mathrm{MSG}}\}),
\label{eq:global_attn_pool}
\end{equation}
where \(\mathcal{V}_{\mathrm{MSG}}\) denotes the set of nodes in the MSG, including ordinary nodes and the super node. 

\(\mathbf{Z}_g^\text{MSG}\) is then fed into the decoder for translation. The encoder remains frozen and the decoder learns a balanced representation to generate translations based on the gate mechanism (see Eq.~\eqref{eq:gate}) between multimodal representations $\mathbf{Z}_g^\text{MSG}$ and the source embedding \( \mathbf{H} \).

\noindent\textbf{Stage~2: Cross-modal Generalization via LSG.}  
As shown in Figure~\ref{fig:framework}~(Left), Stage~2 inputs the same GAT adapter with LSGs built from texts, allowing multimodal knowledge to be cross-modally generalized to broader image-free domains.
In the LSG, each ordinary node \(i\) represents a textual entity with the initial embedding \(\mathbf{Z}_i^{(0)}\) and a super node of the global text embedding, \(\mathbf{Z}_{\mathrm{SN}}^{\mathrm{LSG}(0)}\) by the M-CLIP text encoder. The embedding \(\mathbf{Z}_i^{(l)}\) of node $i$ at the layer \(l\) for an ordinary node is:
\begin{equation}
\begin{split}
\mathbf{Z}_i^{(l)}=\sigma(\!\!\!\!
\sum_{j\in\mathcal{N}_i^{\mathrm{LSG}}}\!\!\!\!\!\alpha_{ij}^{(l)}\phi(\mathbf{W}[\mathbf{Z}_i^{(l-1)}\|\mathbf{Z}_j^{(l-1)}\|\mathbf{E}_{ij}])\\
\quad+\alpha_{i,\mathrm{SN}}^{(l)}\phi(\mathbf{W}[\mathbf{Z}_i^{(l-1)}\|\mathbf{Z}_{\mathrm{SN}}^{\mathrm{LSG}(l-1)}\|\mathbf{E}_{i,\mathrm{SN}}])
).
\end{split}
\label{eq:ordinary_node_lsg_l}
\end{equation}
The LSG super node is updated as:
\begin{equation}
\mathbf{Z}_{\mathrm{SN}}^{\mathrm{LSG}(l)}\!\!=\!\sigma(\!\!\!\!
\sum_{i\in\mathcal{N}_{\mathrm{SN}}^{\mathrm{LSG}}}\!\!\!\!\!\alpha_{\mathrm{SN},i}^{(l)}\phi(\mathbf{W}[\mathbf{Z}_{\mathrm{SN}}^{(l-1)}\|\mathbf{Z}_i^{(l-1)}\|\mathbf{E}_{\mathrm{SN},i}])
),
\label{eq:supernode_lsg_l}
\end{equation}
with $\mathcal{N}_{\mathrm{SN}}^{\mathrm{LSG}}$ denoting all LSG ordinary nodes. The LSGs help the generalization of shared multimodal knowledge weight $\mathbf{W}$ from $\mathcal{D}_m$ in Stage~1 to the image-free domain $\mathcal{D}_l$ in Stage~2.

Similar to Eq.~\eqref{eq:global_attn_pool}, we obtain the graph representation of LSG $\mathbf{Z}_g^{\mathrm{LSG}}$. The mBART decoder is enhanced by generalized knowledge $\mathbf{Z}_g^{\mathrm{LSG}}$ from $\mathcal{D}_m$ and adapted to the image-free domain $\mathcal{D}_l$ with unfrozen mBART encoder hidden state.

\subsubsection{Cross-modal GAT Adapter}
We employ a multi-layer GAT with residual connections~\cite{he_deep_2015,velickovic_graph_2018,li_deepgcns_2019} to learn multimodal knowledge and cross-modally generalize it to a broader image-free domain. Figure~\ref{fig:framework}~(Right) shows the architecture of the GAT adapter fusing the output from the mBART encoder and enhancing the mBART decoder. We denote \(\mathbf{Z}_g\) as the graph representation of MSG or LSG by the Global Attention Pooling. The fusion output $\mathbf{O}$ between the graph representation \(\mathbf{Z}_g\) and mBART encoder hidden states \(\mathbf{H}\) is performed via the cross attention as:
\begin{equation}
    \mathbf{A} = \text{MultiHeadAttn}(\mathbf{H}, \mathbf{Z}_g, \mathbf{Z}_g) + \mathbf{H}
\end{equation}
\begin{equation}
    \mathbf{O} = \text{LayerNorm}(\text{Dropout}(\mathbf{A}))
\end{equation}
A gate mechanism $\mathbf{g}$ balances the embedding flow:
\begin{equation}
    \mathbf{g} = \sigma(\mathbf{W}_2\text{ReLU}(\mathbf{W}_1[\mathbf{O}\| \mathbf{H}]))
    \label{eq:gate}
\end{equation}
\begin{equation}
    \mathbf{H}' = \text{LayerNorm}(\mathbf{g} \odot \mathbf{O} + (1-\mathbf{g}) \odot \mathbf{H})
\end{equation}
where \(\odot\) denotes element-wise multiplication and \([\cdot\|\cdot]\) represents concatenation, $ \mathbf{H}'$ is the input of mBART decoder. This two-stage process demonstrates the central inductive role of the cross-modal GAT adapter in representing and generalizing structured multimodal relationships.

\begin{table*}[t]
\centering
\resizebox{\textwidth}{!}{
\begin{tabular}{l|ccc|ccc|c}
\hline
\textbf{Model} & \multicolumn{3}{c|}{\textbf{EN $\rightarrow$ DE}} & \multicolumn{3}{c|}{\textbf{EN $\rightarrow$ FR}} & \textbf{Mean $\Delta$} \\
\cline{2-7}
 & Test2016 & Test2017 & MSCOCO & Test2016 & Test2017 & MSCOCO & \\
\hline
 mBART (NMT backbone)~\cite{liu_multilingual_2020}& 41.12 & 36.63 & 32.89 & 63.37 & 57.01 & 47.28 & -2.08\\
\hline
\multicolumn{8}{c}{\textbf{MMT Model with Multimodal Inference}}\\
\hline
 DCCN~\cite{lin_dynamic_2020}                        & 39.70 & 31.00 & 26.70 & 61.20 & 54.30 & 45.40 & -5.41\\
 GMNMT~\cite{yin_novel_2020}                         & 39.80 & 32.20 & 28.70 & 60.90 & 53.90 & -     & -5.14\\
 Gated Fusion*~\cite{wu_good_2021}                  & 42.00 & 33.60 & 29.00 & 61.70 & 54.80 & 44.90 & -4.13\\
 WRA-MNMT~\cite{wra_papers}                & 39.30 & 32.30 & 28.50 & 61.70 & 54.10 & 43.40 & -5.02\\
 $\text{UMMT}^\#$~\cite{fei_scene_2023}         & 37.40 &   -   &   -   & 56.90 &  -    &  -    & -7.68\\
 Soul-Mix~\cite{cheng_soul-mix_2024}                 & \textbf{44.24} & 37.14 & 34.26 & 64.75& 57.47 & 49.25 & -0.61\\
\hline
 \textbf{GIIFT~(with image)}                 &       43.32&       \underline{37.47}&       \underline{34.66}&       \underline{65.17}&       \textbf{59.11}&       \textbf{49.76}&       -0.21\\
\hline
\multicolumn{8}{c}{\textbf{MMT Model with Image-free Inference}}\\
\hline
 ImagiT~\cite{long_generative_2021}                  & 38.50 & 32.10 & 28.70 & 59.70 & 52.40 & 45.30 & -5.68\\
 VALHALLA~\cite{li_valhalla_2022}                    & 41.90 & 34.00 & 30.30 & 62.20 & 55.10 & 45.70 & -3.58\\
 VALHALLA*~\cite{li_valhalla_2022}                   & 42.70 & 35.10 & 30.70 & 63.10 & 56.00 & 46.50 & -2.78\\
 UMMT~\cite{fei_scene_2023}                          & 32.00 &   -   &   -   & 50.60 &   -   &   -   & -13.53\\
 CLIPTrans~\cite{gupta_cliptrans_2023}               & 43.87 & 37.22& 34.49 & 64.55 & 57.59 & 48.83 & -0.7\\
\hline
 \textbf{GIIFT~(image-free)}                               & \underline{44.04} & \textbf{38.41} & \textbf{34.94} & \textbf{65.61} & \underline{58.05} & \underline{49.72} &  \\
\hline
\end{tabular}
}
\caption{BLEU on the Multi30K.  $\Delta$ is gap vs. ``GIIFT~(image-free)''. ``GIIFT~(with image)'' is trained and tested with images and texts in only one stage with unfrozen mBART. * represent ensembled models, \# denotes the model trained and tested with images and texts.}
\label{tab:mmt_models}
\end{table*}
\begin{table*}[t]
\centering
\resizebox{\textwidth}{!}{
\begin{tabular}{l|ccc|ccc|c}
\hline
\textbf{Model} & \multicolumn{3}{c|}{\textbf{EN $\rightarrow$ DE}} & \multicolumn{3}{c|}{\textbf{EN $\rightarrow$ FR}} & \textbf{Mean $\Delta$} \\
\cline{2-7}
 & Test2016 & Test2017 & MSCOCO & Test2016 & Test2017 & MSCOCO & \\
\hline
mBART (NMT backbone) & 69.59 & 65.07 & 60.15 & 82.40 & 77.63 & 71.58 & -1.35\\
\hline
\multicolumn{8}{c}{\textbf{MMT Model with Multimodal Inference}}\\
\hline
DCCN~\cite{lin_dynamic_2020}                     & 56.80 & 49.90 & 45.70 & 76.40 & 70.30 & 65.00 & -11.73\\
GMNMT~\cite{yin_novel_2020}                       & 57.60 & 51.90 & 47.60 & 74.90 & 69.30 & - & -9.72\\
Gated Fusion*~\cite{wu_good_2021}          & 67.80 & 61.90 & 56.10 & 81.00 & 76.30 & 70.50 & -3.48\\
WRA-MNMT~\cite{wra_papers}                & 58.30 & 52.80 & 48.50 & 76.30 & 70.60 & 63.80 & -8.92\\
$\text{UMMT}^\#$~\cite{fei_scene_2023}         & 57.20 &  -   &  -   & 70.70 &  -    &  -    & -13.42\\
Soul-Mix~\cite{cheng_soul-mix_2024}          & 69.93 & 63.59 & 59.94 & 83.24 & 78.23 & 73.48 & -1.01\\
\hline
\textbf{GIIFT~(with image)}           &       \underline{70.65}&       \underline{65.59}&       \underline{61.37}&       \underline{83.32}&       \textbf{78.95}&       \textbf{73.98}&       -0.10\\
\hline
\multicolumn{8}{c}{\textbf{MMT Model with Image-free Inference}}\\
\hline
ImagiT~\cite{long_generative_2021}              & 55.70 & 52.40 & 48.80 & 74.00 & 68.30 & 65.00 & -11.72\\
VALHALLA~\cite{li_valhalla_2022}         & 68.80 & 62.50 & 57.00 & 81.40 & 76.40 & 70.90 & -2.92\\
VALHALLA*~\cite{li_valhalla_2022}             & 69.30 & 62.80 & 57.50 & 81.80 & 77.10 & 71.40 & -2.43\\
UMMT~\cite{fei_scene_2023}                  & 52.30 &  -   &  -   & 64.70 &  -    &  -    & -18.87\\
CLIPTrans~\cite{gupta_cliptrans_2023}         & 70.22 & 65.43 & 61.26 & 82.48 & 77.82 & 72.78 & -0.75\\
\hline
\textbf{GIIFT~(image-free)}                             & \textbf{71.08} & \textbf{65.88} & \textbf{61.66} & \textbf{83.65} & \underline{78.36} & \underline{73.86} &  \\
\hline
\end{tabular}
}
\caption{METEOR on the Multi30K. $\Delta$ is gap vs. ``GIIFT~(image-free)''. ``GIIFT~(with image)'' is trained and tested with images and texts in only one stage with unfrozen mBART. * represent ensembled models, \# denotes the model trained and tested with images and texts. }
\label{tab:meteor}
\end{table*}

\begin{table}[htbp]
\centering
\resizebox{0.5\textwidth}{!}{
\begin{tabular}{l|cc|cc|cc}
\hline
\multirow{2}{*}{\textbf{Model}} 
  & \multicolumn{6}{c}{\textbf{EN $\rightarrow$ UK}} \\ 
\cline{2-7}
  & \multicolumn{2}{c|}{Test2016} 
  & \multicolumn{2}{c|}{Test2017} 
  & \multicolumn{2}{c}{MSCOCO} \\ 
\cline{2-7}
  & BLEU & METEOR & BLEU & METEOR & BLEU & METEOR \\
\hline

mBART             & 53.43 & 74.67 & 46.05 & 68.86 & 44.05 & 66.75 \\
CLIPTrans
                  &   54.01&   74.76&   46.30&   69.23&   44.11&   66.87\\
\textbf{GIIFT~(image-free)}& \textbf{55.10} &\textbf{ 75.18} & \textbf{47.85 }& \textbf{70.01 }& \textbf{44.93} & \textbf{67.05} \\
\hline
\end{tabular}
}
\caption{BLEU and METEOR on EN $\rightarrow$ UK (Ukraine) task of the Multi30K.}
\label{tab:low_resource}
\end{table}

\begin{table}[htbp]
\centering
\resizebox{0.5\textwidth}{!}{
\begin{tabular}{l|cc|cc}
\hline
\textbf{Model} & \multicolumn{2}{c|}{\textbf{EN $\rightarrow$ DE}} & \multicolumn{2}{c}{\textbf{EN $\rightarrow$ FR}} \\
\cline{2-5}
 & BLEU & METEOR & BLEU & METEOR \\
\hline
mBART  & 15.58 & 41.18 & 26.50 & 52.06 \\
\hline
CLIPTrans & 16.63 & 42.13 & 26.78 & 51.76 \\
\hspace*{2em}(\textit{w/o.} Stage~1) & 17.60 & 42.81 & 27.71 & 53.38 \\
\hline
\textbf{GIIFT~(image-free)} & \textbf{18.10} & \textbf{43.88} & \textbf{28.70} & \textbf{54.58} \\
\hspace*{2em}(\textit{w/o.} Stage~1) & \underline{17.79}& \underline{43.01}& \underline{27.89}& \underline{53.45}\\
\hline
\end{tabular}
}
\caption{Comparison of domain generalization on text-only WMT. ``(\textit{w/o.} Stage~1)'' denotes model trained without image involvement from Multi30K.}
\label{tab:wmt_result}
\end{table}

\section{Experiments}
\noindent\textbf{Datasets.} We conduct experiments on two benchmarks: Multi30K~\cite{elliott_multi30k_2016} and WMT2014~\cite{bojar_findings_2014}. Multi30K is a widely used MMT benchmark as a multilingual extension of the Flickr30k. WMT is a text-only multilingual NMT dataset. For evaluation, we conduct EN$\rightarrow$\{DE, FR\} translation tasks on Multi30K's three standard test splits: Test2016, Test2017, and MSCOCO. 
We also train and test EN$\rightarrow$\{DE, FR\} on WMT with multimodal knowledge from Multi30K to evaluate the inductive image-free inference. We downsample WMT train set matching Multi30K's size, while keeping the validation and test sets unchanged.\\
\noindent\textbf{Implementation Details.} We train GIIFT on an A100 GPU, with AdamW optimizer (polynomial decay). The batch size is 64, learning rate is \(2e^{-5}\) (Stage~1) and \(1e^{-5}\) (Stage~2). The GAT Adapter is 9 layers with the same 1024 dimensions as M-CLIP. Text decoding is beam search with size 5. Implementations are in PyTorch and Huggingface Transformers library. We report BLEU~\cite{papineni_bleu_2001} and METEOR via SacreBLEU~\cite{post_call_2018} and the evaluate library~\cite{banerjee_meteor_2005}, respectively. Results are three-run averages with early-stop patience 5 on BLEU. We chose mBART as our backbone to ensure a fair comparison with baselines such as CLIPTrans~\cite{gupta_cliptrans_2023} and to highlight our improvements. All tables bold the best and underline the second-best. Baseline figures are derived from papers or repositories. All the numbers of each model reported are from their fine-tuned version on the corresponding dataset.

\subsection{Results on Multi30K}

Table~\ref{tab:mmt_models} and~\ref{tab:meteor} contain translation performance comparison of BLEU and METEOR on EN$\rightarrow$\{DE, FR\} tasks of Multi30K. 
Table~\ref{tab:low_resource} shows the BLEU and METEOR on EN $\rightarrow$ UK (Ukraine) task of Multi30K compared with baselines. ``GIIFT~(with image)'' is a single-stage variant where both training and testing use paired images and texts, and the mBART backbone remains fully unfrozen throughout. ``GIIFT (image-free)'' is the full image-free model trained in two-stage framework as Subsection~\ref{sec:giift}. 

From Table~\ref{tab:mmt_models} and~\ref{tab:meteor}, we observe that GIIFT~(image-free) surpass the strongest baseline, Soul-Mix (even tested with images), with an average gain of +0.61~(1.37\%) BLEU and +1.01~(1.55\%) METEOR, and over the best scene-graph-based image-free baseline, UMMT, by +13.525~(42.27\%) BLEU and +18.865~(36.07\%) METEOR. 
It shows the effectiveness of GIIFT by preserving the entire information from images and texts.

Moreover, the GIIFT~(image-free) model with cross-modal generalizing multimodal knowledge for image-free inference and GIIFT~(with image) with multimodal inference secure top-two ranks on 5 out of 6 benchmarks with overall parity. GIIFT~(image-free) even surpasses GIIFT~(with image) by +0.21~(0.63\%) BLEU and +0.1~(0.17\%) METEOR on average. In contrast, UMMT degrades dramatically without images, scoring on average -5.4~(-12.76\%) BLEU and -5.45~(-8.53\%) METEOR below $\text{UMMT}^{\#}$ with multimodal inference.
This demonstrates that the GIIFT has the robustness of cross-modal generalization for image-free inference, which can mitigate the gap between multimodal inference and image-free inference.

From Table~\ref{tab:low_resource}, we can find that GIIFT~(image-free) consistently outperforms both mBART and CLIPTrans baselines on the EN$\rightarrow$\{UK\} task of Multi30K. We can observe that GIIFT achieves an average of +1.45 higher than mBART and +1.15 higher than CLIPTrans in BLEU; and an average of +0.65 higher than mBART and +0.46 higher than CLIPTrans in METEOR. 
It shows the effectiveness of GIIFT by preserving and learning multimodal information from images and texts for cross-modal generalization.

\subsection{Results on WMT}

In Table~\ref{tab:mmt_models}, CLIPTrans outperforms other image-free inference baselines, so we adopt it as our primary baseline and use its official repository for experiments. Like GIIFT~(image-free), this two-stage mBART-based model is trained with Stage~1 on Multi30K and Stage~2 on WMT. The results of CLIPTrans and GIIFT in Table~\ref{tab:wmt_result} are obtained under the same model parameter setup. 

Table~\ref{tab:wmt_result} shows that GIIFT achieves the highest BLEU and METEOR scores overall, while also significantly outperforming GIIFT~(\textit{w/o.} Stage~1) trained without images from Multi30K on both metrics. 
This validates GIIFT's inductive ability to achieve robust image-free inference via cross-modal generalization.

The difference between CLIPTrans and GIIFT stems from how each model handles the modality gap and the resulting impact on cross-modal generalization.
CLIPTrans attempts to align images to textual captions for transferring the Stage 1 visual features into Stage 2 caption translations via a mapping network, which limits the multimodal correlation for cross-modal generalization.
By contrast, GIIFT can effectively embrace the modality gap via graph-guided fusion and achieve inductive generalization via a cross-modal GAT adapter. In Stage 1, all the modalities are learned and represented in a \textit{unified multimodal knowledge space} via multimodal scene graphs~(MSGs). In Stage 2, the proposed cross-modal GAT adapter generalizes that knowledge into the text-only domain via the assistance of linguistic scene graphs~(LSGs). Benefit from a two-stage inductive learning via MSGs or LSGs, GIIFT shows better performance in achieving robust image-free inference performance.

\begin{table}[htbp]
  \centering
  \resizebox{0.48\textwidth}{!}{%
  \begin{tabular}{lcccc}
    \hline
    \textbf{Model}        & \textbf{BLEU} & \textbf{Completeness} $\uparrow$ & \textbf{Ambiguity} $\downarrow$ & \textbf{Fluency} $\uparrow$ \\
    \hline
    mBART                & 63.37         & 7.0            & 7.3         & 8.1         \\
    CLIPTrans            & 64.55         & 8.0            & 5.5         & 8.8         \\
    GIIFT~(image-free)              & \textbf{65.61}         & \textbf{9.3}             & \textbf{4.6}          & \textbf{9.5}          \\
    \hspace*{1em}(\textit{w/o.} Stage~1)   & \underline{64.91}          & \underline{8.8}             & \underline{5.1}          & \underline{9.0}          \\
    \hline
  \end{tabular}
  }
  \caption{Human evaluation metrics (10-point Likert scale) and BLEU scores on Multi30K EN$\to$FR task.}
  \label{tab:human_eval}
\end{table}

\subsection{Human Evaluation}

Table~\ref{tab:human_eval} presents human evaluation on the Multi30K EN$\to$FR test set using a 10-point Likert scale for completeness, ambiguity, and fluency, alongside BLEU scores for reference.

Results in Table~\ref{tab:human_eval} show that our GIIFT substantially enhances key translation dimensions that correlate with human satisfaction. Compared with the baselines, GIIFT significantly outperforms in completeness, indicating that the model can utilize more multimodal context information. GIIFT achieves the highest completeness, fluency ratings, and the lowest ambiguity among all baselines, demonstrating practically meaningful gains even when BLEU is already high. These gains highlight that our method delivers practical improvements across dimensions that BLEU alone cannot capture.

\subsection{Comparison with Multimodal LLMs}

\begin{table*}[h]
\centering
\resizebox{\textwidth}{!}{
\begin{tabular}{l|cc|cc|cc|cc|cc|cc}
\hline
\textbf{Model} & \multicolumn{6}{c|}{\textbf{EN $\rightarrow$ DE}} & \multicolumn{6}{c}{\textbf{EN $\rightarrow$ FR}} \\
\cline{2-13}
 & \multicolumn{2}{c|}{Test2016} & \multicolumn{2}{c|}{Test2017} & \multicolumn{2}{c|}{MSCOCO} & \multicolumn{2}{c|}{Test2016} & \multicolumn{2}{c|}{Test2017} & \multicolumn{2}{c}{MSCOCO} \\
\cline{2-13}
 & BLEU & METEOR & BLEU & METEOR & BLEU & METEOR & BLEU & METEOR & BLEU & METEOR & BLEU & METEOR \\
\hline
LlaVA-7B& 27.15 & 58.54 & 23.70 & 52.05 & 19.54 & 47.77 & 35.67 & 65.57 & 34.79 & 62.94 & 35.00 & 62.87 \\
LlaVA-34B& 25.30 & 58.76 & 25.16 & 55.58 & 22.04 & 51.53 & 40.25 & 69.47 & 38.95 & 67.36 & 39.99 & 68.82 \\
 Llama3-70B& 40.49& 69.09& 36.12& 65.06& 34.38& 60.86& 50.95& 77.13& 49.39& 74.54& 49.38&73.39\\
mBART          & 41.12 & 69.59 & 36.63 & 65.07 & 32.89 & 60.15 & 63.37 & 82.40 & 57.01 & 77.63 & 47.28 & 71.58 \\
GIIFT~(image-free)      & \textbf{44.04} & \textbf{71.08} & \textbf{38.41} & \textbf{65.88} & \textbf{34.94} & \textbf{61.66} & \textbf{65.61} & \textbf{83.65} & \textbf{58.05} & \textbf{78.36} & \textbf{49.72} & \textbf{73.86} \\
\hline
\end{tabular}
}
\caption{Comparison with Multimodal LLM on Multi30K.}
\label{tab:llm_comparison}
\end{table*}

Table~\ref{tab:llm_comparison} shows the experimental comparison of BLEU and METEOR on EN$\rightarrow$\{DE, FR\} tasks of Multi30K with multimodal LLMs. 

We use an RTX A6000 GPU to employ LlaVA-7B, LlaVA-34B~\cite{liu_visual_2023} and Llama3-70B~\cite{grattafiori2024llama3herdmodels} based on Ollama. We adopt LLaVA’s few-shot inference paradigm~\cite{10.5555/3495724.3495883} by prompting the model with a small set of (image, source caption, target translation) examples drawn from Multi30K and then asking it to translate a new image’s caption into the target language. 

From Table~\ref{tab:llm_comparison}, we observe that our GIIFT achieves the highest BLEU and METEOR scores on EN$\rightarrow$\{DE, FR\} benchmarks, with significant improvements over mBART and LLMs. These results confirm the effectiveness of our GIIFT model and show that compact, domain-specialized multimodal MMT models can outperform larger general-purpose LLMs on the translation task.

Considering the parameter complexity compared with multimodal LLMs, the GIIFT only introduced the GAT adapter in the framework, which is lightweight and efficient for training. For instance, the mBART backbone has approximately 600M parameters, while our GIIFT framework adds only about 33.2M parameters in total, approximately 27M in the GAT adapter layers and 6.2M in the gated fusion layer. 
Although LLaVA has far more parameters for supporting broad capabilities, our GIIFT’s parameters are wholly devoted to yielding greater efficiency and accuracy for the translation.

\begin{table*}[htbp]
\centering
\resizebox{\textwidth}{!}{
\begin{tabular}{l|cc|cc|cc|cc|cc|cc}
\hline
 \multirow{3}{*}{\textbf{Model}} & \multicolumn{6}{c|}{\textbf{EN $\rightarrow$ DE}} & \multicolumn{6}{c}{\textbf{EN $\rightarrow$ FR}} \\
\cline{2-13}
 & \multicolumn{2}{c|}{Test2016} & \multicolumn{2}{c|}{Test2017} & \multicolumn{2}{c|}{MSCOCO} & \multicolumn{2}{c|}{Test2016} & \multicolumn{2}{c|}{Test2017} & \multicolumn{2}{c}{MSCOCO} \\
\cline{2-13}
 & BLEU & METEOR & BLEU & METEOR & BLEU & METEOR & BLEU & METEOR & BLEU & METEOR & BLEU & METEOR \\
\hline
mBART (backbone) & 41.12 & 69.59 & 36.63 & 65.07 & 32.89 & 60.15 & 63.37 & 82.40 & 57.01 & 77.63 & 47.28 & 71.58 \\
\hline
GIIFT~(\textit{w/o.} Stage~1)& 
\underline{43.63} & \underline{70.95} & 
37.76 & \underline{65.49} & 
\underline{34.47} & \underline{60.81} & 
\underline{64.91}& \underline{83.07}& 
\underline{57.71}& \underline{78.01}& 
\underline{48.95}& \underline{73.20}\\
 GIIFT~(\textit{w/o.} freezing)& 42.84& 70.37& 37.24& 65.35& 34.38& 60.69& 63.74& 82.62& 56.52& 77.23& 48.92&72.54\\
GIIFT~(\textit{w/o.} gate) & 
43.50 & \underline{70.95} & 
\underline{37.96} & 65.11 & 
33.85 & 60.21 & 
64.14 & 82.61 & 
57.58 & 78.00 & 
48.59 & 72.90 \\
\hline
\textbf{GIIFT~(image-free)} & 
\textbf{44.04} & \textbf{71.08} & 
\textbf{38.41} & \textbf{65.88} & 
\textbf{34.94} & \textbf{61.66} & 
\textbf{65.61} & \textbf{83.65} & 
\textbf{58.05} & \textbf{78.36}& 
\textbf{49.72} & \textbf{73.86} \\
\hline
\end{tabular}
}
\caption{Ablation study on Multi30K. ``GIIFT~(\textit{w/o.} freezing)'' has an unfrozen mBART encoder in Stage~1. ``GIIFT~(\textit{w/o}. Stage~1)'' is trained without images. ``GIIFT~(\textit{w/o}. gate)'' is trained in two stages without gate fusion.}
\label{tab:ablation_study}
\end{table*}

\subsection{Ablation Study}
To further verify the effectiveness of the different components in the GIIFT framework, we showed the experimental results of the ablated versions in
Table~\ref{tab:ablation_study} on EN$\rightarrow$\{DE, FR\} Multi30K. 

As shown in Table~\ref {tab:ablation_study}, the GIIFT~(image-free) with two-stage continuous learning by the proposed GAT adapter achieves the best performance across all benchmarks. Removing the gating mechanism, GIIFT~(\textit{w/o.} gate), results in a more significant reduction in BLEU and METEOR scores, underscoring the critical balancing role of the gating mechanism to fuse multimodal knowledge and mBART information. Additionally, omitting the multimodal learning stage 1, GIIFT~(\textit{w/o.} Stage~1), leads to a decrease in performance, highlighting the importance of learning generalizable multimodal knowledge from MSGs. The performance of GIIFT~(unfrozen) is significantly lower than GIIFT~(image-free), and closer to the backbone model mBART. This underscores freezing the mBART encoder in Stage~1 to maintain its stable embedding. Compared with the backbone mBART, there is a substantial drop in BLEU and METEOR, which demonstrates the effectiveness of the GIIFT framework in learning multimodal knowledge and cross-modal generalization for image-free inference.

\begin{figure*}[ht!]
    \centering
    \includegraphics[width=1\linewidth]{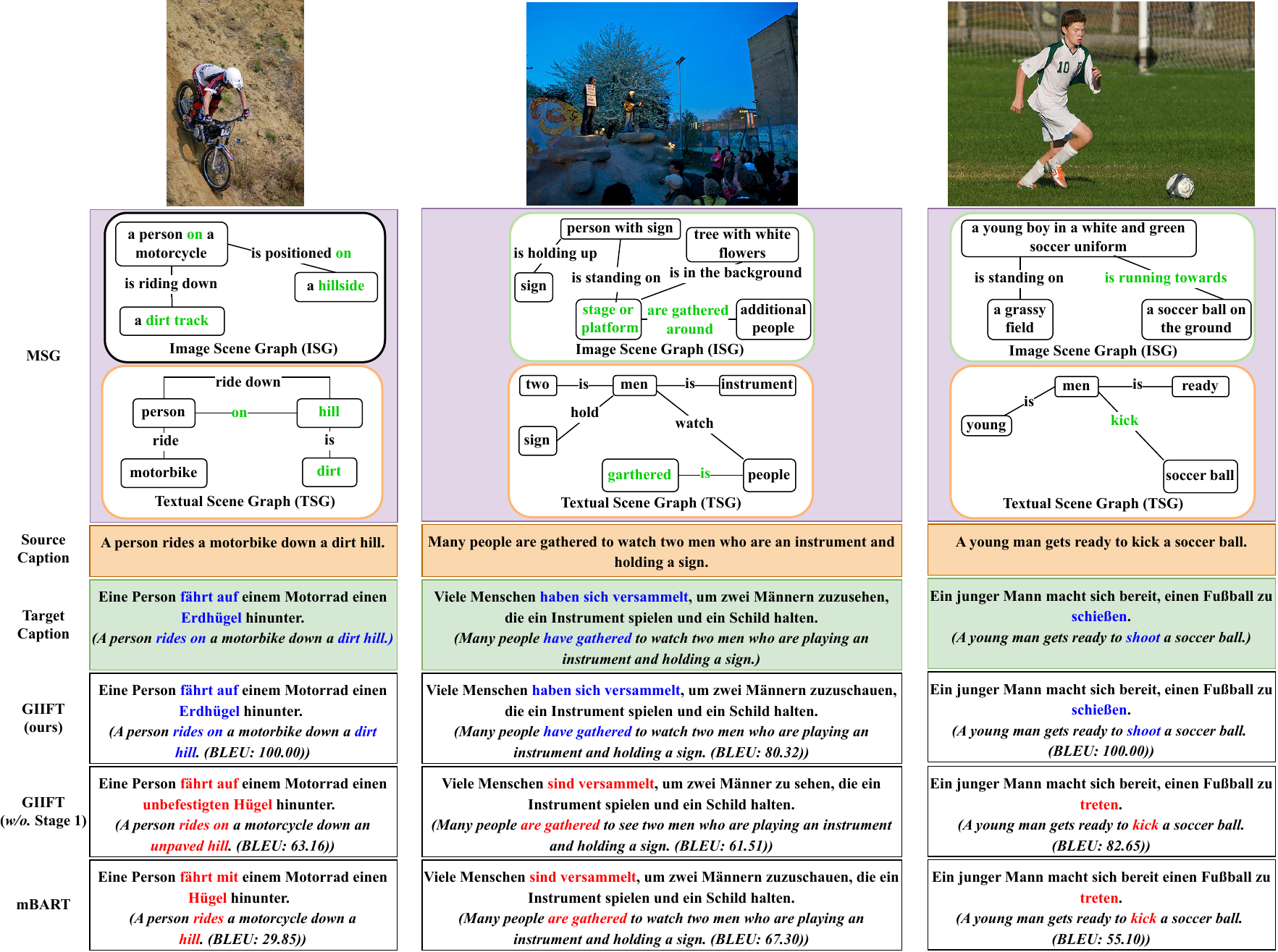}
    \caption{
    Under image-free inference, full GIIFT~(image-free) is compared to GIIFT~(\textit{w/o.} Stage~1) and mBART on Multi30K validation set. The italicized bracketed translations of the German caption mark the differences in red.
}
    \label{fig:case}
\end{figure*}

\section{Case Study}

To investigate the advantages of GIIFT, we examine cases in comparing: the GIIFT~(image-free), which learns multimodal knowledge from MSGs for cross-modal image-free generalization via LSGs; GIIFT~(\textit{w/o.} Stage~1), which only learns linguistic relationships from LSGs; and the text-only mBART.

\textbf{(i) Environmental scene.} 
In Figure~\ref{fig:case}~(left), GIIFT~(image-free) and GIIFT~(\textit{w/o.} Stage~1) both correctly capture the spatial preposition ``on'' through MSG or LSG, respectively, despite the source English caption omitting the spatial information. But lacking the scene graph, mBART produces imprecise translations without ``on'', which highlights the functions of LSGs to guide the learning and generalize the spatial relation knowledge. This case shows that LSGs guide GIIFT to better generalize spatial relation information from MSGs’ multimodal knowledge in Stage~2 for image-free translation inference.
Additionally, as shown in Figure~\ref{fig:case}~(left), the environmental scene ``dirt hill'', is only accurately translated by full GIIFT~(image-free) with multimodal knowledge from MSGs. GIIFT~(\textit{w/o.} Stage~1) and mBART, although ``dirt'' in the source caption, produce imprecise translations, reflecting overlooking of the modality-specific information. This case shows the effectiveness of MSGs that can well embrace modality gaps by multimodal graph fusion and fully preserve multimodal information for improving translation. Other cases from Test2016 are shown in Appendix~\ref{appendix:case}.

\textbf{(ii) Temporal states.} In Figure~\ref{fig:case}~(middle), the MSG captures a scene with ``are gathered around the stage or platform'', enabling full GIIFT to recognize it as a completed state and generate the appropriate perfect tense in German. In contrast, both GIIFT~(\textit{w/o.} Stage~1) and mBART, limited by modality gap, cannot capture the temporal state from an aligned visual-linguistic space, which eliminates the unique temporal state from the image. So they translate English ``are gathered'' directly into the German present tense.

\textbf{(iii) Action states.} Figure~\ref{fig:case}~(right) shows the MSG's action state ``is running towards'' guides GIIFT~(image-free) to correctly translate the action as ``shoot'' rather than ``kick''. The visual information, available only through MSG, enables the correct translation from multimodal knowledge. GIIFT~(\textit{w/o.} Stage~1) and mBART, however, can only literally translate to ``kick'' in German, which also shows the importance of accepting different modality-specific information rather than alignment.

\section{Conclusion}

This work introduces GIIFT, a two-stage graph-guided inductive MMT framework, along with novel Multimodal and Linguistic Scene Graphs. GIIFT outperforms existing MMT models on Multi30K even with image-free inference, demonstrating the effectiveness of learning multimodal knowledge in a unified space via MSGs and achieving cross-modal generalization via LSGs. Its image-free performance remains robust, matching its multimodal-inference counterpart. GIIFT also outperforms both the text-only NMT backbone and leading image-free MMT baselines on WMT, showing effective induction of multimodal knowledge to broader text-only domains. 
Further analysis highlights the advantages of multimodal graph-guided generalization for image-free inference and confirms the effectiveness of the two-stage framework with a lightweight but efficient GAT adapter for cross-modal inductive learning.


\clearpage 

\appendix

\section{Scene Graph Prompts for LLaVA}\label{sec:prompt}
\begin{tcolorbox}[width=\textwidth, 
    before skip=2pt,   %
    after skip=0pt,    %
    top=3pt,          %
    bottom=3pt,
    left=3pt,
    right=3pt,
    breakable
    ]
{\small
Please analyze the image provided and construct a structured scene graph, adhering to the following guidelines, and represent it in a JSONL (JSON Lines) format:

1. Entities: List all significant objects or subjects visible in the image, which may include things, animals, or people. Describe each entity in detail, noting their quantities, colors, and any distinctive features. Each description should be distinct and consistent across the document to ensure clarity.\\
2. Relations: Define all pivotal relationships between the entities using tuples. Each tuple must maintain the exact terminology used in the entities' descriptions. These relationships should be expressed as triplets: [subject entity, predicate, object entity]. Importantly, ensure that the scene graph forms a connected structure. Every entity appearing as a subject or object in one relation must connect to another entity in a different relation, preventing any isolated nodes or subgraphs within the graph. In cases involving an entity related to multiple others, such as being 'between' or 'consist of' them, express this by dividing the relationship into distinct tuples using descriptors like 'is positioned between' and 'and also between' to maintain clarity. Generate triplets with a subject, an active verb or relational word, and a distinct object. Each triplet should clearly describe an action or relationship, avoiding states or implied conditions.\\
Avoid focusing on too detailed or minor elements that do not significantly contribute to the scene's overall understanding. Use active verbs that show a clear action or relationship. Avoid state or possession verbs like "have" that imply a condition without a distinct action. Incorrect Relations Examples to Avoid:\\

1.["one person in red shirt", "one dog", "one cat"] (lacks clear action)\\
2.…………\\
Correct Relations Examples of the above, the number of the example is the same as the number of the incorrect example:\\

1.["one person in red shirt", "is holding", "a book"]\\
2.…………\\
Key Point: Ensure every triplet uses an active verb or distinct relational word to connect the subject and object, clearly describing a specific action or relationship and forming a triplet.\\

This structure ensures that the scene graph is comprehensive and interconnected, accurately reflecting the dynamics and layout of the scene. The response must strictly follow the JSONL format specified here and not include any extraneous text.\\
 
This is a scene graph JSONL example response of the Example Image, the entity\_descriptions1, entity\_descriptions2, entity\_descriptions3, entity\_descriptions4 and entity\_descriptions5 need to be replaced by specific entities in the image. The relation word1, relation word2, and relation word3 are also need to be replaced by the specific action or relation you observe in the given image. Also, the number of entities and relations is not fixed. It should depend on the given image. The following scene graph JSONL is just an example. You need to describe the real relations based on your given image.\\

\texttt{\{"entities": ["entity\_descriptions1", "entity\_descriptions2", "entity\_descriptions3", "entity\_descriptions4", entity\_descriptions5], "relations": [["entity\_descriptions1", "relation word1", "entity\_descriptions3"], ["entity\_descriptions2", "relation word2", "entity\_descriptions4"], ["entity\_descriptions1", "relation word3", "entity\_descriptions5"]]\}}\\

You must not include the word 'image' in the scene graph JSONL. You must not copy the example above! You must describe the entities and their relationships in the given image. Now, you must respond to the scene graph based on the image provided! Straitly follow my instructions. Now what is the scene graph of the image?
}
\end{tcolorbox}

\noindent We use an RTX A6000 GPU to employ Llava-34B. For each query, the system info explains the task description and provides guidelines for scene graph generation in JSONL format. We also have a quality assessment script to discard any ISG data that fails to meet our generation task description presented in the prompts. We take a temperature of 0 as the default for multimodal large language models (MLLMs) to have a relatively robust performance. If the MLLM can’t generate scene graphs that meet the requirements with a temperature of 0, we’ll switch to a temperature of 0.4. We exclusively use MLLM for image preprocessing to generate scene graphs according to our task description, with quality ensured by the script.

\clearpage
\begin{figure}[ht]

    \centering
    \includegraphics[width=2\linewidth]{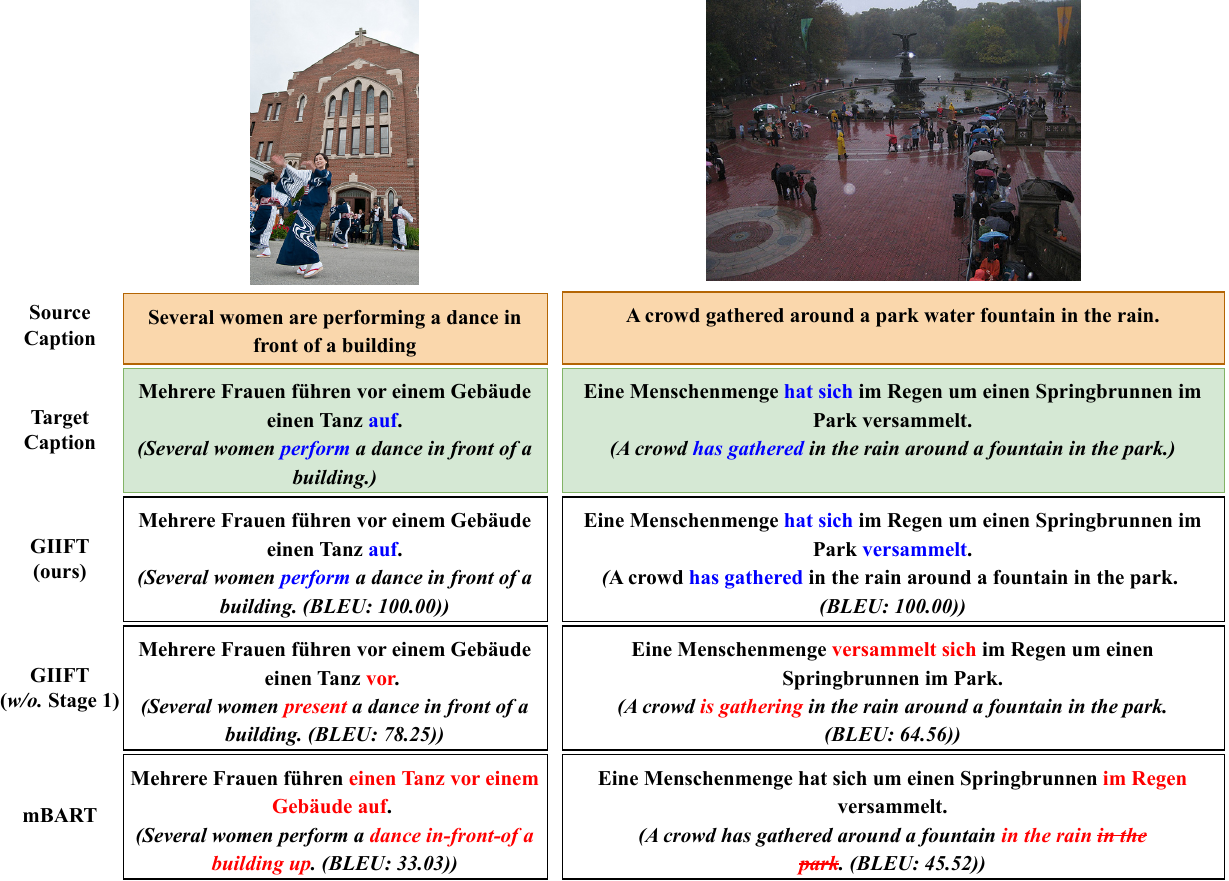}
    \parbox{\textwidth}{  
\caption{Case study of GIIFT on image-free inference when compared to GIIFT~(\textit{w/o.} Stage~1) and the mBART. Data points are drawn from the Test2016 set of Multi30K. The gold sentence represents the ground truth. The italicised sentence in the bracket presents the English translation of the German text, while red words highlight the crucial translation differences.}
\label{fig:2016case}
}    
\end{figure}
\section{Case Study on Multi30K Test2016 Set}\label{appendix:case}
As shown in Figure~\ref{fig:2016case} (Left), our full model, GIIFT~(image-free), correctly associates the text with visual scene context to translate the separable verb accurately, using ``auf'' to express the ``perform'' action, whilst GIIFT~(\textit{w/o.} Stage~1) incorrectly translates it as ``vor'' to express ``present''. mBART fails to properly understand the scene, resulting in disordered word arrangement.

As shown in Figure~\ref{fig:2016case}~(Right), our full model, GIIFT~(image-free), correctly generalizes the tense from textual to visual information, accurately translating the crowd's gathered state as ``has gathered'' rather than directly translating ``gathered''. GIIFT~(\textit{w/o.} Stage~1) incorrectly translates it as the progressive tense ``is gathered''. Although mBART uses the correct tense, it still fails to properly understand the scene context, omitting the location ``in the park'' and misattributing the modifier ``in the rain'', leading to overall semantic confusion.

\end{document}